%% file: main.tex
\documentclass[runningheads]{llncs}

\usepackage{eccv}

\usepackage{eccvabbrv}

\usepackage{graphicx}
\usepackage{booktabs}
\usepackage{graphicx}
\usepackage{amsmath}
\usepackage{amssymb}
\usepackage{booktabs}
\usepackage{multirow}
\usepackage{xspace}
\usepackage{xcolor}
\usepackage{subcaption}

\usepackage[accsupp]{axessibility}  
\newcommand{\papername}{ADen\xspace}

\newcommand{\image}{\mathbf{I}}
\newcommand{\pose}{\mathbf{P}}

\usepackage{hyperref}

\usepackage{orcidlink}

\begin{document}

\title{\papername: Adaptive Density Representations for Sparse-view Camera Pose Estimation}
\titlerunning{\papername}

\author{Hao Tang \and
Weiyao Wang \and Pierre Gleize \and Matt Feiszli}

\authorrunning{H. Tang et al.}

\institute{FAIR at Meta \\
\email{\{haotang, weiyaowang, gleize, mdf\}@meta.com}
}

\maketitle

\input{sections/abstract}

\input{sections/introduction}

\input{sections/related_work}

\input{sections/method}

\input{sections/experiments}

\input{sections/conclusion}

\clearpage  

%
%
\bibliographystyle{splncs04}
\bibliography{egbib}
\end{document}




\title{\papername: Supplementary Material}

\author{Hao Tang \and
Weiyao Wang \and Pierre Gleize \and Matt Feiszli}

\authorrunning{H. Tang et al.}

\institute{FAIR at Meta \\
\email{\{haotang, weiyaowang, gleize, mdf\}@meta.com}
}
\maketitle

\section{MLE for KDE and mixture models}
As mentioned in Section 3 that in our setup, the pose generator produces samples from the current estimate for the pose distribution.  While we design our system as a generator-discriminator pair with a contrastive loss, we observe that if the set of samples is sufficient to approximate the distribution, then this is in fact simply a maximum-likelihood objective.  This suggests classical models like kernel density estimates and mixture models may also be effective here. We first introduce KDE and MM and subsequently show an ablation study using these methods.

For the KDE and MM variants, the objective is again maximum likelihood, but we now have an explicit formula for the density.  For KDE, 
\begin{align*}
    p_{\mathrm{KDE}}(P) = \frac{1}{Nh^d} \sum_{n=1}^N k\left(\frac{\mathrm{d}(P, P_n)}{h} \right)
\end{align*}
where $k$ is the exponential kernel $k(x) = \exp(-x)$, $\mathrm{d}$ is geodesic distance, and $h$ is a fixed kernel bandwidth hyperparameter.  The density for a mixture model is similar, but adds data-dependent, per-component weight and scale parameters
\begin{align*}
    p_{\mathrm{MM}}(P) = \sum_{n=1}^N \frac{w_n}{h_n^d} f\left(\frac{\mathrm{d}(P, P_n)}{h_n} \right)
\end{align*}
where $f$ is a non-normalized density and the weights are positive and satisfy $\sum w_n = 1$.  Both the mixture weights and and scale $h_n$ are produced by the generator.  We use the exponential density $f(x) = \exp(-x)$.

In both cases we use the generator's output to define the distribution, and minimize the negative log-likelihood (via the explicit formula) at the ground truth location in pose space.  In these settings, there is no discriminator; during training we optimize the likelihood directly and at inference time we select the predicted location with the greatest density.

\textbf{Contrastive Loss vs Maximum Likelihood}
Let $\hat p$ be our contrastive likelihood defined above.  We may also define a traditional density 
\begin{align}
    \hat q(P | \image_1, ..., \image_N, i) = \frac{1}{Z}h(\pose)
\label{eq-generic-likelihood}
\end{align}
\noindent where $h(\pose) = \exp(x(P))$ is our non-normalized density and
the normalizing constant $Z$ is obtained by integrating over pose space.  This integral can be tricky; if the space is large (e.g. the space of natural images) computing $Z$ can be prohibitive.  However, if we can obtain a representative set of samples from the density, then the next well-known \cite{carreira2005contrastive} result shows we can obtain the gradient of $Z$ even in high-dimensional spaces (see appendix for detail.):
\begin{lemma}
With notation as above, 
\begin{align}
\label{eq-grad-partition}
    \nabla \log Z &= \int_\mathcal{P} \nabla x(\pose) \hat q(\pose) d\pose
\end{align}
\end{lemma}


\noindent where pose space $\mathcal{P}$ is the manifold $\SO$ for rotation, or $\SO \times \mathbb{R}^3$ for rotation and translation.  Since our generator is explicitly designed to produce samples $\samp \sim \hat q$, we can approximate (\ref{eq-grad-partition}) as
\begin{align}
    \nabla \log Z &\approx \frac{1}{N} \sum_{n=1}^N \nabla x(\pose_n)
    \label{eq-approximate-sampled-gradient}
\end{align}
In other words, if our samples accurately represent the model distribution, we can compute a maximum-likelihood estimate by combining (\ref{eq-generic-likelihood}) and (\ref{eq-approximate-sampled-gradient}):
\begin{align}
\log \hat q(x^*) \approx x^* - \frac{1}{N} \sum_{n=1}^N \nabla x(\pose_n)
\end{align}
which is closely related to our contrastive loss.

\textbf{Experiments with KDE and MM on samples}
In this ablation, we first use KDE or MM on the generated samples to show they indeed work as well as using the discriminator to find the mode. Then we show an ablation where we train the generator using the KDE or MM loss. As observed in \cref{tab:kde_mm_results}, we notice an important drop in performance when training those two variants from scratch (using a standard negative log likelihood loss). 
This suggests KDE and MM are both families of model that can reach top-performance on this problem, but suffer from optimization difficulties. We leave the exploration of those challenges as future work. To produce those results, we hand-tuned the kernel bandwidth to 0.13 (correspond to a $2\sigma$ angle error of 15 degrees), and in the case of MM, the kernel weights are obtained from the discriminator's softmax output. The maximum likelihood estimates (MLE) are obtained by returning the generated sample having the largest probability. In the discriminator's case, that probability is given by its softmax output, in the KDE / MM case, it is obtained by computing their analytical density.

\begin{table}[]
\centering
\begin{tabular}{c | lccc}
\hline
Setting             &                                     & Acc@5 & Acc@10 & Acc@15 \\ \hline
\multicolumn{3}{l}{\textbf{Inference using KDE or MM}}                                          \\ \hline
                    & Discriminator & 55.7 & 79.9 & 86.4 \\
                    & KDE          & 54.3 & 79.5 & 86.4  \\
                    & MM           & 54.0 & 79.3 & 86.1  \\ \hline
\multicolumn{6}{l}{\textbf{Training using KDE or MM}}                                    \\ \hline
\multirow{2}{*}     & KDE & 38.6 & 69.9 & 80.5 \\
                    & MM & 43.7 & 74.0 & 83.00 \\ \hline

\end{tabular}
\caption{\textbf{Ablation on using KDE or MM.} First, we observe that swapping the inference method of the normally trained model achieves similar results. This suggests both KDE and MM could potentially converge to a similar solution. Second, when trained from scratch using a NLL loss, both KDE and MM suffer from a performance drop, highlighting the training difficulty when using these models.}
\label{tab:kde_mm_results}
\vspace{-6mm}
\end{table}

\newpage
{\small
\bibliographystyle{ieee_fullname}
\bibliography{egbib}
}

%% file: sections/abstract.tex
\begin{abstract}

Recovering camera poses from a set of images is a foundational task in 3D computer vision, which powers key applications such as 3D scene/object reconstructions. Classic methods often depend on feature correspondence, such as keypoints, which require the input images to have large overlap and small viewpoint changes.  Such requirements present considerable challenges in scenarios with sparse views. Recent data-driven approaches aim to directly output camera poses, either through regressing the 6DoF camera poses or formulating rotation as a probability distribution. However, each approach has its limitations. On one hand, directly regressing the camera poses can be ill-posed, since it assumes a single mode, which is not true under symmetry and leads to sub-optimal solutions. On the other hand, probabilistic approaches are capable of modeling the symmetry ambiguity, yet they sample the entire space of rotation uniformly by brute-force.  This leads to an inevitable trade-off between high sample density, which improves model precision, and sample efficiency that determines the runtime. In this paper, we propose \papername to unify the two frameworks by employing a generator and a discriminator: the generator is trained to output multiple hypotheses of 6DoF camera pose to represent a distribution and handle multi-mode ambiguity, and the discriminator is trained to identify the hypothesis that best explains the data. This allows \papername to combine the best of both worlds, achieving substantially higher precision as well as lower runtime than previous methods in empirical evaluations.

\end{abstract}

%% file: sections/introduction.tex
\section{Introduction}
\label{sec:intro}
Understanding 3D structure from 2D image observations of objects or scenes is an important task in computer vision. Recent advances in Neural Radiance Field (NeRF)~\cite{mildenhall2021nerf} and Gaussian Splatting~\cite{kerbl3Dgaussians} enable high quality 3D reconstruction and novel view synthesis, initially from densely posed images. At the core of these methods, Structure-from-Motion (SfM) plays an important role to extract camera poses from the input images.

Motivated by key real-world applications such as online marketplaces and casual captures by everyday users~\cite{zhang2022relpose}, there is also growing interest in bringing these methods to sparse-view images~\cite{niemeyer2022regnerf,reizenstein2021common,yu2021pixelnerf}, where only a handful of images (\eg 3 to 5) are available, each covers a different viewpoint. Similar to their dense-view counterparts, these methods often assume known input camera poses; yet in practice, the default geometric-based SfM pipeline may fail due to minimal view overlap in the context of sparse-view images. This highlights the need for sparse pose estimations and inspires a new stream of research taking data-driven approach that learns to predict poses from large-scale object-centric dataset with superior performance than geometric-based approach~\cite{zhang2022relpose,lin2023relpose++,sinha2023sparsepose,wang2023posediffusion,murphy2021implicit}.



\begin{figure}[t]
\centering
\includegraphics[width=0.9\columnwidth]{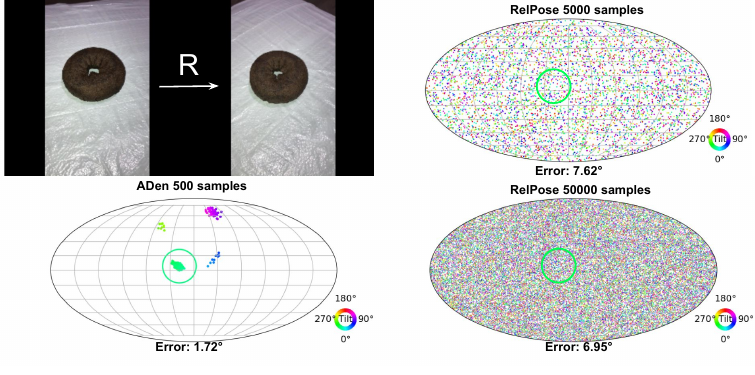}
\caption{\textbf{Ambiguity in wide baseline images}. Implicit-PDF/RelPose models rotation as a probability distribution using an energy-based method, which requires evaluating densely sampled rotation hypotheses. To achieve high accuracy, RelPose requires assessing 500k rotations for each image pair, incurring significant computational costs. In contrast, \papername outputs 500 high accuracy hypotheses directly, avoiding the constraints imposed by grid resolution. Filled circles are samples while unfilled circles are the ground truth relative rotation. 
}
\label{fig:motivation}
\vspace{-6mm}
\end{figure}

The data-driven approaches for recovering relative camera poses from sparse-view images can be broadly classified into two categories. One approach is to directly regress the 6DoF camera parameters rotation $R$ and translation $T$ \cite{kendall2015posenet,sinha2023sparsepose}. However, in the sparse view setting, ambiguity arises, parituclarly for objects or scenes with symmetry (\Cref{fig:motivation}). Pose regression assumes a single mode in the data, which can result in suboptimal solutions when trained with data exhibiting a multi-mode distribution.

A different approach is to model the rotation as a probability distribution \cite{murphy2021implicit,lin2023relpose++,zhang2022relpose}. Implicit-PDF \cite{murphy2021implicit} first introduces a method to predict arbitrary, non-parametric probability distributions over the rotation manifold to address the symmetry issue. It densely samples rotation hypotheses from SO(3), and predicts the probability for each given image features. It naturally accounts for the uncertainty in the symmetric case, allowing the model to output multiple modes, thus improving accuracy over pose regression. 

While powerful, the primary drawback of this brute-force energy-based approach lies in the requirement to densely sample from the entire parameter space. To achieve high accuracy, a dense grid must be sampled from the parameter space. For example, RelPose \cite{zhang2022relpose} needs to sample 500,000 rotation matrices at inference time to generate a dense enough grid for good accuracy, in particular at lower error thresholds. However, evaluating 500,000 rotation hypotheses for one pair of images is computationally expensive. Furthermore, this method suffers from the curse of dimensionality: it is only practical in low dimensions (\eg SO(3) is three dimensional) and becomes prohibitively expensive in higher dimensions. Moving from 3 to 6 dimensions to jointly model rotation and translation~\cite{murphy2021implicit} without reducing sampling granularity implies a 250-billion-sized grid, yet reducing the sampling granularity implies inferior results.

How do we benefit from both approaches? Our key observation is that using uniform grid to represent pose distributions is inefficient. In real-world, the distribution of poses is highly skewed, with a few isolated modes dominating the distribution. In other words, the real-world distribution of poses sit in between the regression (one mode)~\cite{kendall2015posenet,sinha2023sparsepose} and the uniform distribution~\cite{murphy2021implicit,lin2023relpose++,zhang2022relpose}. Based on this observation, 
we focus on a generator-discriminator framework: given 2 or more images, the generator learns to produce samples from the conditional distribution of relative poses, while the discriminator ranks them.  We find that this requires only a few hundred samples to cover all possible modes of the distribution. The adaptive nature also eliminates any fundamental lower bound on accuracy imposed by a grid resolution (either a fixed spacing for fixed grids, or expected spacing for random grids).

We name this work \papername; as shown in \Cref{fig:motivation}, \papername only needs 500 samples to outperform methods that sample 500K locations at inference time; \papername is not constrained by any grid resolution and can output samples arbitrarily close to the true mode. The generated samples clearly learn to follow the multi-modal distribution, capturing the uncertainty of poses.  Importantly, by eliminating the need for dense sampling from the parameter space, \papername is not limited to model rotation alone; the generator easily outputs joint rotation and translation [$\mathbf{R}$, $\mathbf{t}$] pairs simply by predicting particles in the product space, \textit{without} increasing their number.

In summary, our contributions are as follows.
\begin{itemize}
    \item We propose \papername, a method for learning and sampling from the conditional distribution of relative pose from images using an efficient, adaptive generator-discriminator framework.
    \item \papername extends naturally to high-dimensional spaces without requiring exhaustive sampling of the entire space; it adapts to the complexity of the distribution and not the ambient space.
    \item Experiments show \papername outperforms SoTA methods by a large margin, especially at low error thresholds. Moreover, \papername runs much faster than previous methods, achieving real-time inference speed. 

\end{itemize}

%% file: sections/related_work.tex
\section{Related work}
\label{sec:related_work}
\subsection{Structure-from-Motion (SfM)}
SfM aims at recovering 3D geometry and camera poses from multi-view images set. This classic problems has been extensively studied in the past \cite{hartley2003multiple, ozyecsil2017survey}, which typically includes computing image features (typically key points) \cite{bay2006surf,lowe1999object}, establishing feature correspondence \cite{lucas1981iterative}, computing camera poses using five-point or eight-point algorithms \cite{hartley2003multiple,hartley1997defense,li2006five} with RANSAC \cite{fischler1981random}, and verifying epipolar geometry with bundle adjustments \cite{triggs2000bundle}. Each component in this pipeline highly depends their previous steps and needs careful tuning to be robust to scale to hundreds or thousands of images \cite{furukawa2010towards,sarlin2019coarse}. Among these algorithms, COLMAP \cite{schonberger2016structure} is an open-source implementation of this entire pipeline and has been widely used by the community. 

Recent progress in deep neural nets and large-scale image datasets, various methods have been proposed to improve the performance of different components of the SfM pipeline with better feature descriptors \cite{detone2018superpoint,dusmanu2019d2,revaud2019r2d2,gleize2023silk}, improved feature matching \cite{sun2021loftr,truong2020glu,sarlin2020superglue}. Although SfM works well with multiple views, it often fails in the wide baseline setting , due to insufficient overlap between images, occlusion and failure to establish correspondence.

\subsection{Data-driven pose estimation}
The key step in the above-mentioned SfM pipeline is to establish correspondence between images. However, given only sparse views or wide baselines, it struggles to find matches \cite{choi2015robust}. As the result of advances in network architecture and availability of large-scale dataset with posed images \cite{reizenstein2021common,downs2022google}, learning-based methods has been proposed recently to directly estimate camera poses between images using a top-down approach. Unlike previous bottom-up approach, PoseNet \cite{kendall2015posenet} uses a network to directly regress the 6DoF camera parameters from a single RGB image. Implicit-PDF \cite{murphy2021implicit} is the first to propose a new framework for single image pose estimation which represents the rotation distribution implicitly, with a neural network to assign probability given input images and a candidate pose and demonstrated strong performance in handling symmetry. RelPose \cite{zhang2022relpose} finds the widespread presence of symmetry in real-world dataset under sparse-view setting and extends the Implicit-PDF framework to estimate camera rotations based on intial pairwise prediction. RelPose++ \cite{lin2023relpose++} further extends RelPose by adding a transformer to process multi-view images jointly and also report camera translation by defining a new camera coordinate system. SparsePose \cite{sinha2023sparsepose} proposes to learn an iterative refinement step to regress initial pose estimation. PoseDiff \cite{wang2023posediffusion} proposes to formulate the SfM problem as a probabilistic diffusion framework which mainly mirrors the iterative procedure of bundle adjustment. 

Similar to PoseNet and SparsePose, our approach involves the direct regression of poses to the ground truth. However, we distinguish our method by employing a generator that produces multiple pose hypotheses. Our strategy differs from the conventional approach of regressing all predictions to a single ground truth. Instead, we utilize losses designed to guide the mode towards the ground truth, leveraging the diversity provided by the generator.
This approach enables the model to explore various modes in the dataset without being constrained to learn a single mode. Unlike Implicit-PDF and RelPose which require evaluating hundreds of thousands of pose hypotheses, \papername only generates a few hundred hypotheses and uses a discriminator to pick the best one thus can achieve real-time inference.

\subsection{Generative learning and contrastive learning}

Modeling pose distribution using generative models has been successful recently \cite{wang2023posediffusion, chen2022epro}. GANs\cite{goodfellow2020generative} are generative models trained adversarially. Similar to GANs, we do use a generator and a discriminator, but with a shared backbone (as opposed to having two fully independent models). Unlike GANs, our objective is not adversarial. The discriminator is trained to score various samples, not competing with the generator.
The discriminator is trained using the contrastive loss \cite{chen2020simple} to model the pose distribution, by rewarding the positive samples and penalizing negative samples. Unlike previous methods \cite{he2022masked,chen2020simple}, we do not aim to learn visual representations but to score samples from a distribution.  





%% file: sections/method.tex
\section{Method}
\label{sec:method}

\begin{figure}[t]
\centering
\includegraphics[width=\textwidth]{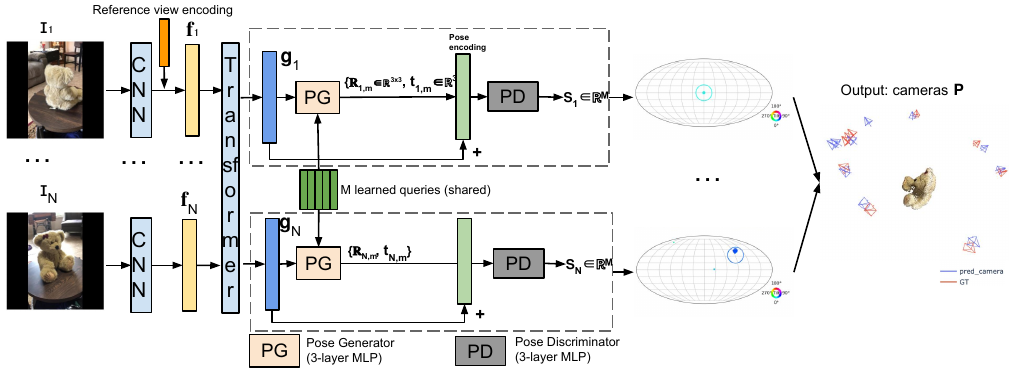}
\caption{\textbf{\papername overview.} \papername is a novel method for recovering camera poses from sparse-view RGB images. \papername starts by extracting per-image features using the ResNet backbone, then uses a transformer to fuse features from all images and propagate information globally. \papername predicts a non-uniform distribution over camera poses for each image by first applying a pose generator head on fused features to produce a support set of $M$ camera poses, then using a pose discriminator with fused features to predict probability on each generated pose. 
}
\label{fig:model}
\vspace{-4mm}
\end{figure}

Given a set of N sparse-view images of an object \{$\mathbf{I}_1$, ..., $\mathbf{I}_N$\}, \papername aims to recover the camera extrinsics \{[$\mathbf{R}_1$, $\mathbf{t}_1$], ..., [$\mathbf{R}_N$, $\mathbf{t}_N$]\}, where $\mathbf{R}_i \in \mathrm{SO}(3)$ is the rotation and $\mathbf{t}_i \in \mathbb{R}^3$ is the translation for image $\mathbf{I}_i$. 
The model is trained to first draw samples $[\mathbf{R}_{n,k}, \mathbf{t}_{n,k}], $ from the distribution of all possible poses given the input images $p(\mathbf{R}_n, \mathbf{t}_n | \mathbf{I}_1, ..., \mathbf{I}_N)$, where n is the index for the image and k is the index for the sample. And subsequently, it scores these samples and selects the most probable one as its prediction $[\hat{\mathbf{R}}_{n}, \hat{\mathbf{t}}_{n}]$ for each image.
Our proposed method consists of a shared feature backbone, and two heads: pose generator and pose discriminator (\Cref{fig:model}). We use the camera pose for the first frame as canonical reference to infer camera poses of other frames. 

\subsection{Multi-view feature extraction}
\papername begins by extracting per-image features using ResNet~\cite{he2016deep}, denoted as $\mathbf{F}_i=f(\mathbf{I}_i) \in \mathbb{R}^d$, where $d$ is the feature space dimension. To reason relative poses across multiple input views, \papername concatenates the features and uses transformers~\cite{vaswani2017attention} $t$ to fuse them: $(\mathbf{G_1}, ..., \mathbf{G}_N)=t(\mathbf{F_1}, ..., \mathbf{F_N})$. An additional reference image encoding is added to $\mathbf{F_1}$ to designate it as representing the canonical pose. The fused features $\mathbf{G}$ are then fed into a pose generator and a pose discriminator to predict relative poses.


\subsection{Pose generator}
\label{sec:pose_generator}
The pose generator takes inputs from the fused features and predicts multiple pose hypotheses for each input image. These hypotheses aim at capturing a distribution over poses under natural ambiguities such as rotation symmetry. However, the training data only contains a single groundtruth pose. Training all pose proposals to match the ground truth will result in a mode collapse:
the generator will output only one pose and loses its ability to output a pose distribution to model these natural ambiguities. This has also been verified through empirical experiments (\Cref{tab:train_discriminator_generator}). To avoid such mode collapse, we allow the model to explore multiple modes by only regressing one pose candidate closest to the ground truth; no loss is computed to penalize other poses predictions.

We use an MLP to generate $M$ pose hypotheses $Q_{i,m} \in \mathbb{R}^{7}$ which contains 4 elements for the quaternion and 3 for the translation. The choice to represent the rotation matrix with a quaternion is due to the ease of normalizing the model's output in quaternion space. We then transform the quaternion representation of the rotation component of $Q_{i,m}$ back to $3\times 3$ rotation matrix. This yields $P_{i,m} \in \mathbb{R}^{12}$.
The pose generator takes inputs from the fused features $G$ and $M$ randomly initialized learnable queries $\left\{ e_{i,m}\right\}_{m=1}^M \in \mathbb{R}^{256}$. An MLP is used to map the queries to the feature space. Note that these randomly initialized queries specialize in distinct partition of the pose space, similar to learnable queries in object detections~\cite{carion2020end}, enabling the model to fully cover the space. 


$$P_{i,m}=\mathbf{MLP}(\mathbf{G}_i + \mathbf{MLP}(e_{i,m}))$$

\subsection{Pose discriminator}
Given the generated pose hypotheses from the pose generator head, the pose discriminator evaluates the probability of each hypothesis being correct. At inference, the hypothesis with the highest probability is used as the output.

The pose discriminator first embeds the generated hypotheses to the same dimension as feature dimension and then uses an MLP to predict logits $x_{i,m}$. 

$$x_{i,m}=\mathbf{MLP}(\mathbf{G}_i + \mathbf{MLP}(P_{i,m}))$$
For any pose and logit pair $(\pose, x)$ we then have probability:

\begin{align}
    \hat p(\pose | \image_1, ..., \image_N; i) = \frac{e^{x}}{e^{x_{i}^*} + \sum_m e^{x_{i,m}}}
    \label{eq-discriminator-prob}
\end{align}

where $(\pose^*_i, x_i^*)$ is the ground truth pose during training. At inference, we omit this term since ground truth is not available.


\subsection{Model training details}
\label{training}




\textbf{Pose generator}
To prevent mode collapse, which occurs when all proposals are trained to match the ground truth as discussed in \Cref{sec:pose_generator}, we only regress the one pose proposal that is closest to the ground truth pose: $P^*_{i}=[\mathbf{R}_{i}^{*}, \mathbf{t}_{i}^{*}]$ and no loss is computed to penalize other poses proposals. We define the "closest" camera pose as the one with the smallest geodesic distance: 
$$\hat{P}_{i}=[\mathbf{\hat{R}}_i, \mathbf{\hat{t}}_{i}]=\arg \min_{P_{i,m}} d(P_{i,m}, P^*_i), m \in [1,\mathrm{M}]$$

The loss function then is the geodesic distance of predicted rotation to ground truth and the L2 loss of the predicted translation.
\vspace{-1mm}
\begin{equation}
\vspace{-1mm}
    \mathcal{L}_g = \sum_i^N\arccos(\mathrm{Tr}(\mathbf{R}_i^{*}\mathbf{\hat{R}}_i)) + \sum_i^N||\mathbf{t}_i^{*} - \mathbf{\hat{t}}_i||
\end{equation}

\textbf{Pose discriminator}
During training, we also add the ground truth to the network so $x_i^*=\mathbf{MLP}(\mathbf{g}_i, \mathbf{MLP}(p_i^*))$. The loss function for the discriminator then is the contrastive negative log-likelihood loss to differentiate ground truth from the rest. 


\vspace{-3mm}
\begin{align}
\vspace{-1mm}
    \mathcal{L}_d &= -\sum_i^N \log \hat p(P_i^* | \image_1, ..., \image_N, i) 
\end{align}
with $\hat p$ given by (\ref{eq-discriminator-prob}).
The final loss is the sum of pose generator loss and pose discriminator loss $\mathcal{L} = \mathcal{L}_g  + \mathcal{L}_d $.

\textbf{Joint training} Training instability is well-studied in generative adversarial models (GAN) \cite{goodfellow2020generative}.  In particular, as the generator improves its ability to generate accurate camera poses, the discriminator's task becomes more difficult, and its gradients may effectively become noise; this can negatively impact model performance. We observed similar behavior during our training.  There are many techniques to address this for GANs, but in our setting a very simple approach works well. Specifically, we add random Gaussian noise into the query embedding $\mathbf{e_i}$ to generate negative examples for the discriminator. In our ablation study, we explore the impact of this noise on the final performance of the model \Cref{sec:ablation}.

In our setup, the pose generator produces samples from the current estimate for the pose distribution.  While we design our system as a generator-discriminator pair with a contrastive loss, we observe that if the set of samples is sufficient to approximate the distribution, then this is in fact simply a maximum-likelihood objective.  
This suggests classical models like kernel density estimates and mixture models may also be effective here, and indeed this is the case (see Supplemental), although our contrastive method provides stronger results.

\subsection{Implementation details}
Since the pose coordinate is ill-defined in an absolute coordinate system, we set the first frame as the canonical frame and set its rotation to identity. A reference frame encoding is added to the first frame to inform the model. For translation, we use a similar object-centric coordinate system as \cite{lin2023relpose++}. We normalize the translation of the canonical frame to be [0, 0, 1] and normalize all translations according to the distance from the canonical camera to the object center. This can be achieved by computing the center of mass of the object point cloud. 

We use the same Anti-alias Res50 backbone as \cite{zhang2022relpose} and a transformer with six layers of multi-head attention. The ResNet backbone is pretrained on ImageNet while the transformer is trained from scratch. We use Adam as the optimizer with learning rate set to 1$e^{-4}$ and the model is trained for 2000 epochs. During training, we randomly sampled different numbers of images ranging from 2-10 for each batch.

%% file: sections/experiments.tex
\section{Experiments}
\label{sec:experiemnts}

\subsection{Experiment setup}
\textbf{Dataset} We use CO3D V2 \cite{reizenstein2021common} for training and testing \papername, which encompasses 51 object-centric sequences.The ground truth pose for each sequence is determined by employing \cite{schonberger2016structure}. In line with \cite{zhang2022relpose,wang2023posediffusion}, \papername is trained on 41 object categories and subsequently evaluated on both the test set of these 41 seen categories and a separate test set of 10 unseen categories. We follow the same test set frame sampling method of \cite{lin2023relpose++} that resamples N images from a sequence 5 times and reports the average as the final accuracy. 

To further evaluate the generalization ability of method, we test \papername on two additional datasets in zero-shot setting (without any finetuning): Objectron \cite{ahmadyan2021objectron} and Niantic Map Free Relocalization (NMFR) \cite{arnold2022map}. Objectron consists of short, object-centric video clips captured for AR use cases; NMFR captures diverse outdoor scenes and structures including sculptures, murals and foundations. Following~\cite{lin2023relpose++}, all methods were trained using CO3D~\cite{reizenstein2021common} and tested zero-shot (no finetuning). For the Objectron dataset, we used the test set of four classes (Camera, Chair, Cup, Shoe), following the same protocol as Relpose++ \cite{lin2023relpose++}. For the Niantic Map Free dataset, we used the validation set, which contains 65 scans. 

\textbf{Evaluation metrics} 
For evaluating the rotation accuracy, we measure the predicted pairwise relative rotation against ground truth rotations and report the proportion of rotation errors that are less than 15 degrees. To show accuracy at a tighter error threshold, we also report rotation errors that are less than 5 and 10 degrees. For evaluating the translation, we follow \cite{lin2023relpose++} to first apply an optimal similarity transform to align the predicted centers with ground truth \cite{umeyama1991least}. We then report the accuracy as the proportion of translation errors that are less than 10\% of the scene scale. The scene scale is determined by the distance from the centroid of all ground truth cameras to the furthest camera.

\subsection{Comparing with SoTA}

\textbf{Baselines} We compare the proposed method with the following state-of-the-art (SoTA) classic correspondence-based and data-driven approaches:

\textit{COLMAP (SP + SG) \cite{schonberger2016structure}}. This represents the SoTA SfM pipeline using the open-source COLMAP implementation, with SuperPoint as key-point features \cite{detone2018superpoint} and SuperGlue for key-point matching \cite{sarlin2020superglue}. 

\textit{SparsePose \cite{sinha2023sparsepose}} SparesePose directly regresses poses and uses an iterative refinement strategy to further refine the initial poses. 

\textit{PoseDiff \cite{wang2023posediffusion}}. PoseDiff formulates pose estimation inside a probabilistic diffusion framework, which mirrors the iterative procedure of bundle adjustment. 

\textit{RelPose and RelPose++ \cite{zhang2022relpose,lin2023relpose++}}. RelPose is the first work that models rotation as a probability distribution and RelPose++ is an extension that adds a transformer module and also reports translation error. RelPose++ represents the SoTA learning-based method on the Co3D dataset. 

\textit{Pose Regression \cite{lin2023relpose++}}. Pose Regression directly regresses poses given sparse view images. We use the one reported in \cite{lin2023relpose++}. 

As shown in \Cref{table:sota_comparison} and \Cref{table:sota_comparison_translation}, \papername achieves SoTA performance on both rotation and translation accuracy compared to previous methods. \papername consistently outperforms other baselines with different numbers of images as input. This performance gain is more prominent at tighter rotation accuracy threshold \Cref{tab:rot_lower_error_threshold}. Compared with Pose Regression and SparsePose, our method allows the model to explore different modes during training to better learn the underlying ambiguous multi-modal distribution, so it performs significantly better. In contrast to RelPose/RelPose++, which also predicts a probabilistic distribution for poses, our approach is not constrained by the resolution of samples from the SO(3) space and can generate samples that closely match the ground truth camera pose. This improvement is more apparent at tighter rotation error thresholds (\Cref{tab:rot_lower_error_threshold}), underscoring the precision of the samples generated by our pose generator. Additionally, our method achieves SoTA results in camera translation errors, primarily due to enhanced rotation accuracy.
On the two additional datasets, \papername demonstrates strong generalization, significantly surpassing prior methods and achieving SOTA performance \Cref{table:objectron} and \Cref{table:nmfr}.

\begin{table}[]
\centering
\begin{tabular}{llcccccccc}

\hline
                        & \# of Images    & 2    & 3    & 4    & 5    & 6    & 7    & 8    \\ \hline
\multirow{6}{*}{\rotatebox[origin=c]{90}{Seen}}   & COLMAP(SP+SG)   & 30.7          & 28.4          & 26.5          & 26.8          & 27.0          & 28.1          & 30.6          \\
                        & RelPose         & 56.0          & 56.5          & 57.0          & 57.2          & 57.2          & 57.3          & 57.2          \\
                        & Pose Regression & 49.1          & 50.7          & 53.0          & 54.6          & 55.7          & 56.1          & 56.5          \\
                        & RelPose++       & 81.8          & 82.8          & 84.1          & 84.7          & 84.9          & 85.3          & 85.5          \\
                        & PoseDiff        & 76.0          & 76.7          & 77.2          & 77.7          & 78.3          & 78.5          & 78.5          \\
& \textbf{Ours}   & \textbf{84.3} & \textbf{85.5} & \textbf{86.0} & \textbf{86.5} & \textbf{87.0} & \textbf{87.1} & \textbf{87.3}
\\ \hline
\multirow{7}{*}{\rotatebox[origin=c]{90}{Unseen}} & COLMAP(SP+SG)   & 34.5          & 31.8          & 31.0          & 31.7          & 32.7          & 35.0          & 38.5          \\
                        & RelPose         & 48.6          & 47.5          & 48.1          & 48.3          & 48.4          & 48.4          & 48.3          \\
                        & Pose Regression & 42.7          & 43.8          & 46.3          & 47.7          & 48.4          & 48.9          & 48.9          \\
                        & SparsePose     &      -         & 65.0          & 68.0          & 70.0          & 67.0          & 72.0          & 72.0          \\
                        & RelPose++       & 69.8          & 71.1          & 71.9          & 72.8          & 73.8          & 74.4          & 74.9          \\
                        & PoseDiff        & 60.3          & 64.0            & 64.9          & 65.6          & 66.4          & 67.1          & 67.5          \\
& \textbf{Ours}     & \textbf{78.6} & \textbf{79.0} & \textbf{80.1} & \textbf{80.9} & \textbf{81.1} & \textbf{81.4} & \textbf{82.1}
\\ \hline
\end{tabular}
\caption{\textbf{Pairwise relative rotation accuracy @ 15.} We measure the relative angular errors between relative predicted and ground truth rotation, and report the accuracy as the proportion of errors less than 15 degrees. \papername consistently outperforms SoTA methods using different number of images as input.}
\label{table:sota_comparison}
\vspace{-4mm}
\end{table}

\begin{table}[]
\vspace{-3mm}
\centering


\begin{tabular}{llcccccccc}
\hline
Acc@5                   & \# of Images  & 2             & 3             & 4             & 5             & 6             & 7             & 8             \\ \hline
\multirow{3}{*}{Seen}   & RelPose++     & 42.1          & 43.6          & 44.4          & 44.7          & 45.0          & 45.1          & 45.4          \\
                        & PoseDiff      & 49.8          & 51.0          & 50.6          & 51.7          & 52.5          & 53.0          & 54.4         \\
& \textbf{Ours} & \textbf{53.5} & \textbf{55.0} & \textbf{55.8} & \textbf{56.0} & \textbf{56.7} & \textbf{56.6} & \textbf{56.9}
                        \\ \hline
\multirow{3}{*}{Unseen} & RelPose++     & 30.7          & 31.9          & 32.8          & 33.5          & 33.9          & 34.0          & 33.9          \\
                        & PoseDiff      & 38.0          & 34.9          & 37.2          & 39.5          & 41.2          & 42.4          & 42.6          \\
                        & \textbf{Ours} & \textbf{45.3} & \textbf{46.4} & \textbf{47.5} & \textbf{48.2} & \textbf{48.2} & \textbf{48.5} & \textbf{48.7}

\\ \hline
\vspace{-4mm}
\end{tabular}

\centering
\begin{tabular}{llccccccccccc}
\hline
Acc@10                  & \# of Images  & 2             & 3             & 4             & 5             & 6             & 7             & 8             \\ \hline
\multirow{3}{*}{Seen}   & RelPose++     & 70.5          & 72.3          & 73.6          & 74.2          & 74.7          & 75.2          & 75.2          \\
                        & PoseDiff      & 68.4          & 70.1          & 70.2          & 71.1          & 72.2          & 72.3          & 72.4          \\
& \textbf{Ours} & \textbf{77.7} & \textbf{79.0} & \textbf{79.4} & \textbf{79.9} & \textbf{80.4} & \textbf{80.5} & \textbf{80.7}
                        \\ \hline
\multirow{3}{*}{Unseen} & RelPose++     & 57.7          & 58.7          & 60.4          & 61.2          & 61.8          & 62.0          & 62.5          \\
                        & PoseDiff      & 50.6          & 55.7          & 57.2          & 57.9          & 58.3          & 60.0          & 60.8          \\
& \textbf{Ours} & \textbf{69.2} & \textbf{70.6} & \textbf{72.0} & \textbf{72.9} & \textbf{73.5} & \textbf{73.7} & \textbf{74.1}
\\ \hline
\end{tabular}
\caption{\textbf{Pairwise relative rotation accuray @ 5$^{\circ}$  and 10$^{\circ}$}. We report Acc@5 and Acc@10 for \papername, PoseDiff and RelPose++. \papername is not constrained by the accuracy of the sampled grid on SO3, thereby achieving greater gains at tighter accuracy thresholds.}
\label{tab:rot_lower_error_threshold}
\vspace{-4mm}
\end{table}

\begin{table}[]
\centering
\begin{tabular}{llccccccc}
\hline
                        & \# of Images    & 3             & 4             & 5             & 6             & 7             & 8             \\ \hline
\multirow{5}{*}{\rotatebox[origin=c]{90}{Seen}}   & COLMAP(SP+SG)   & 35.8          & 26.1          & 21.6          & 18.9          & 18.3          & 19.2          \\
                        & Pose Regression & 87.6          & 81.2          & 77.6          & 75.8          & 74.5          & 73.6          \\
                        & RelPose++       & 92.3          & 89.1          & 87.5          & 86.4          & 85.9          & 85.5          \\
                        & PoseDiff        & 78.3          & 78.1          & 79.1          & 79.4          & 79.4          & 79.4          \\
                        & \textbf{Ours}   & \textbf{92.5} & \textbf{89.4} & \textbf{88.1} & \textbf{86.7} & \textbf{86.4} & \textbf{85.9}
            \\ \hline
\multirow{5}{*}{\rotatebox[origin=c]{90}{Unseen}} & COLMAP(SP+SG)   & 37.9          & 29.3          & 24.7          & 23.1          & 23.5          & 25.3          \\
                        & Pose Regression & 82.8          & 74.0          & 70.0          & 67.8          & 65.8          & 65.3          \\
                        & RelPose++       & 82.5          & 75.6          & 71.9          & 69.9          & 68.5          & 67.5          \\
                        & PoseDiff        & 61.8          & 61.9          & 63.1          & 63.1          & 62.7          & 63.3          \\
                        & \textbf{Ours}   & \textbf{85.4} & \textbf{78.9} & \textbf{75.7} & \textbf{73.0} & \textbf{71.5} & \textbf{70.7}
\\ \hline
\end{tabular}
\caption{\textbf{Translation accuracy @ 0.2} We measure the accuracy as the proportion of predicted camera translations that are within 20\% of the scene scale of each sequence.}
\label{table:sota_comparison_translation}
\vspace{-4mm}
\end{table}

\begin{table}[h]
    \centering
\begin{subtable}[t]{0.49\textwidth}
\centering
\resizebox{\columnwidth}{!}{
\begin{tabular}[t]{lllllll}
\hline
              & \multicolumn{3}{c}{Rotation}                  & \multicolumn{3}{c}{Cam. Cen.}                 \\ 
\# of Images  & \multicolumn{1}{c}{3}             & \multicolumn{1}{c}{5}             & \multicolumn{1}{c}{8}             & \multicolumn{1}{c}{3}             & \multicolumn{1}{c}{5}             & \multicolumn{1}{c}{8}             \\ \hline
MediaPipe \cite{lugaresi2019mediapipe}     & 52.3          & 52.8          & 52.7          & 74.5          & 59.1          & 49.9          \\
PoseDiff \cite{wang2023posediffusion}      & 69.2          & 68.0          & 70.0          & 87.2          & 73.8          & 67.2          \\
RelPose++ \cite{lin2023relpose++}     & 75.8          & 76.6          & 77.0          & 91.6          & 83.9          & 77.6          \\
\textbf{Ours} & \textbf{86.0} & \textbf{85.8} & \textbf{85.7} & \textbf{93.1} & \textbf{85.0} & \textbf{80.0} \\ \hline
\end{tabular}
}
\caption{Objectron \cite{ahmadyan2021objectron}
}. 
\label{table:objectron}
\end{subtable}
    \hfill 
\begin{subtable}[t]{0.49\textwidth}
\centering
\resizebox{\columnwidth}{!}{
\begin{tabular}[t]{lllllll}
\hline
              & \multicolumn{3}{c}{Rotation}                  & \multicolumn{3}{c}{Cam. Cen.}                 \\ 
\# of Images  & \multicolumn{1}{c}{3}             & \multicolumn{1}{c}{5}             & \multicolumn{1}{c}{8}             & \multicolumn{1}{c}{3}             & \multicolumn{1}{c}{5}             & \multicolumn{1}{c}{8}             \\ \hline
PoseDiff \cite{wang2023posediffusion}      & 39.5          & 43.2          & 42.3          & 70.4          & 48.9          & 42.8          \\
RelPose++ \cite{lin2023relpose++}     & 45.2          & 44.8          & 46.2          & 73.9          & 50.4          & 43.2          \\
\textbf{Ours} & \textbf{55.3} & \textbf{52.6} & \textbf{52.8} & \textbf{74.5} & \textbf{51.5} & \textbf{44.4} \\ \hline
\end{tabular}
}
\caption{Niantic Map-Free Relocalization \cite{arnold2022map}}. 
\label{table:nmfr}
\end{subtable}
\vspace{-4mm}
    \caption{\textbf{Zero-shot evaluation on two additional datasets on rotation (@ 15$^{\circ}$) and camera center (@ 0.2) Accuracy.}}
\vspace{-8mm}
\end{table}

\subsection{Ablation}
\label{sec:ablation}

\begin{figure*}[t]
\centering
\includegraphics[width=\linewidth]{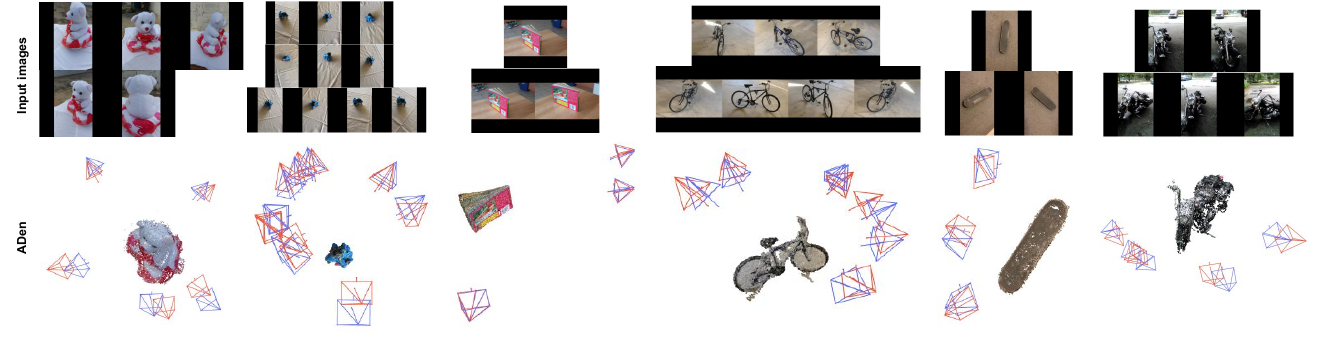}
\caption{\textbf{Camera pose prediction} of \papername on CO3D examples.}
\label{fig:vis_cameras}
\vspace{-4mm}
\end{figure*}

\begin{figure*}[t]
\centering
\includegraphics[width=\linewidth]{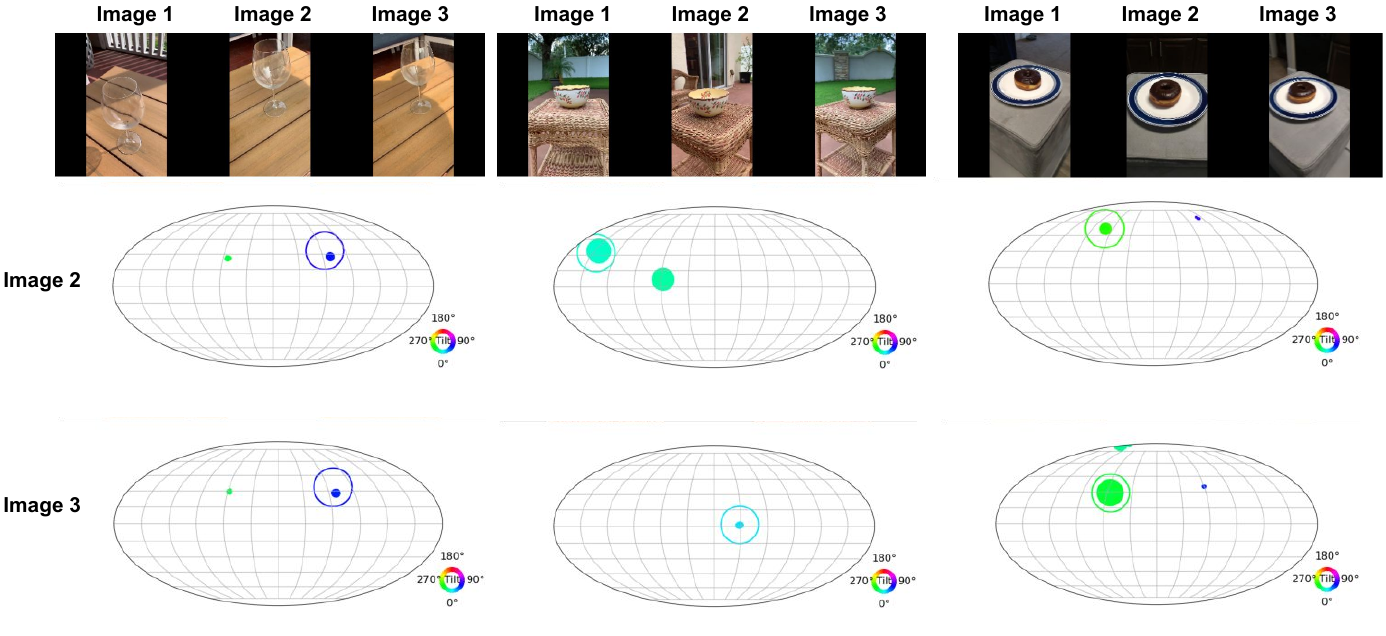}
\caption{\textbf{Relative rotation prediction.} We visualize the relative rotations predicted by \papername on ambiguous cases. The circle size of each filled circle represents the probability assigned by the discriminator. The unfilled larger circle is the ground truth. }
\label{fig:vis_cameras}
\vspace{-6mm}
\end{figure*}

For ablation study, we by default use 5 input images to train and test \papername on the seen categories.

\textbf{How to train the generator.}
\begin{table}[]
\vspace{-6mm}
\centering
\begin{tabular}{c | lcccccc}
\hline

Setting             &                                     & Acc@5 & Acc@10 & Acc@15 \\ \hline
\multicolumn{5}{l}{\textbf{How to train generator}}                                 \\ \hline
                                    & Top 1                               & 55.7      & 79.9      & 86.4      \\
                                    & Top 10                              & 54.5      & 79.8      & 86.8      \\
                                    & Top 50                              & 56.0      & 80.0      & 87.0      \\
                                    & All                                 & 48.2      & 70.5      & 77.3      \\ \hline
\multicolumn{5}{l}{\textbf{How to train discriminator}}                             \\ \hline
\multirow{3}{*}{(1)} & 500 randomly sampled                & 38.2      & 72.0      & 83.4   \\
                   & 5k randomly sampled                 & 50.0      & 78.9      & 87.6   \\
                   & 50k randomly sampled                & 54.2      & 80.2      & 87.4    \\ \hline
\multirow{2}{*}{(2)} & 500 generated w/o noise             & 42.5      & 67.3      & 77.0   \\
                    & 2x training & 48.0      & 71.7      & 79.6   \\ \hline
(3)                  & 500 generated w/ noise              & 55.7      & 79.9      & 86.4

\\ \hline
\end{tabular}
\caption{\textbf{Ablation on training the generator and discriminator.} Training the generator with regression loss applied only to the nearest few samples allows the model to explore alternative modes, leading to improved outcomes. Additionally, training the discriminator with generated noisy camera poses strikes an effective balance between accuracy and training efficiency.}
\label{tab:train_discriminator_generator}
\vspace{-20mm}
\end{table}


Training generator directly controls the quality of the generated sample distribution and thus is critical to the performance of the model. By default we only apply $\mathcal{L}_g$ to the camera pose that is closest to the ground truth. This allows the model to move samples closer to the ground truth, while simultaneously grants the flexibility for other samples to explore different modes without penalizing them. We compare training the generator by pulling 2\% (10), 10\% (50), and all samples towards the ground truth. As shown in \Cref{tab:train_discriminator_generator}, it indicates that applying regression loss to all samples may hinder the model's ability to explore diverse modes during training, leading to 
poorer performance. Furthermore, the generator's sensitivity to the exact number of samples used for training is low, provided there is sufficient leeway for other samples to explore. 

\textbf{How to train the discriminator.}

Training the discriminator is critical to learn good features that are informative in differentiating different camera poses. 
Our observations indicate that the training dynamics of the discriminator, when utilizing samples generated by the pose generator, resemble those observed in generative adversarial networks (GANs) \cite{goodfellow2020generative}, which may lead to instability in training. To mitigate this, we introduce random noise to the generated queries, enhancing the discriminator's training stability. We evaluate different strategies for training the discriminator to ascertain the impact of this added noise: (1) use randomly sampled camera poses; (2) use the generated samples from the pose generator; (3) use a noisy version of the generated samples from the pose generator. For 1) we explore the effects of using 500, 5,000, and 50,000 randomly sampled camera poses for training. For (3), we add Gaussian noise to the learnt queries embedding to produce noisy camera poses for the discriminator $\mathbf{\mathcal{N}(0,3) + e_i}$.

The results are shown in \Cref{tab:train_discriminator_generator}. First, using randomly generated camera poses shows that the denser the sampling, the more improved the outcomes. This improvement can be attributed to denser sampling increasing the probability of the samples being closer to the true mode, which in turn provides hard negatives that facilitate the learning process of the discriminator. Second, training directly with the camera poses generated by the model poses a challenge; as the generator becomes more adept at producing accurate poses, the task of distinguishing between them becomes increasingly difficult for the discriminator, potentially causing training instability. This could potentially lead to instability in training. An attempt to mitigate this by doubling the training duration resulted in a 5\% increase in performance on the Acc@5 metric. However, this still falls short when compared to other training strategies, underscoring the inherent difficulties in training the discriminator. Lastly, introducing a noisy version of the generated samples to the discriminator's training regimen led to an approximate 2-3\% improvement across most accuracy thresholds, achieving results comparable to those obtained by training with a large pool of randomly sampled poses (50k in the first setting). This suggests that adding noise to the generated samples can effectively enhance discriminator performance, paralleling the benefits of extensive random sampling.


\begin{figure}[]
\vspace{-8mm}

\begin{subfigure}{0.49\linewidth}
\centering
\includegraphics[width=\linewidth]{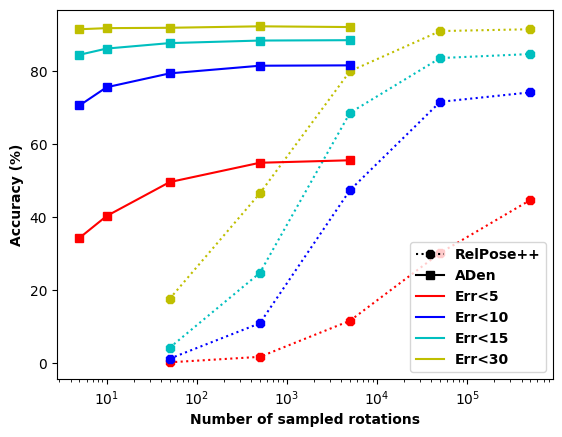}
\caption{\textbf{Rotation accuracy with different number of sampled camera poses (N=5).} 
The proposed method, which directly generates samples, achieves high accuracy with just a few hundred samples. In contrast, RelPose++ relies on random rotations sampled uniformly from SO(3), necessitating a larger sample size to adequately cover the parameter space. This difference becomes more pronounced when measuring accuracy at lower error thresholds.}
\label{fig:diff_n_samples_perf}
\end{subfigure}%
~
\begin{subfigure}{0.49\linewidth}
\centering
\includegraphics[width=\linewidth]{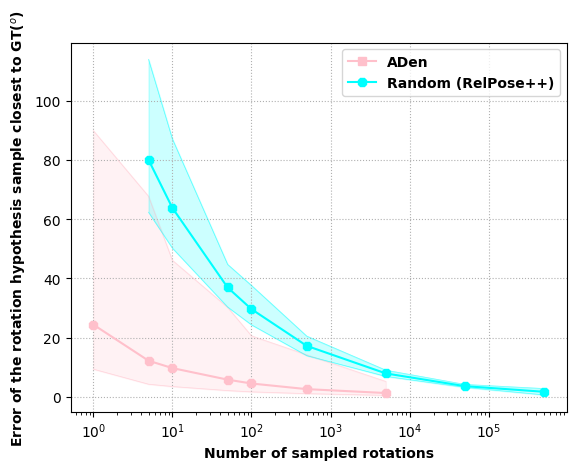}
\caption{\textbf{Generator performance measured as the error of the rotation hypothesis that is closest to the GT.} 
We compare our generated samples with those randomly sampled from SO(3). We assess our generator's effectiveness using the error from the closest rotation hypothesis to the ground truth, as the true pose distribution is unknown. While dense sampling from SO(3) might require over 500,000 samples to cover all rotations, our generator achieves similar accuracy with just 500 samples.
}
\label{fig:generator_num_samples}
\end{subfigure}%
\caption{\textbf{Performance of the generator and discriminator as number of sampled poses.}}
\vspace{-6mm}
\end{figure}
We demonstrate the performance variations with different numbers of samples, primarily focusing on a comparison with RelPose++ because it also draws samples during inference. The key distinction lies in our method's utilization of the generator to produce samples, as opposed to RelPose++ which randomly selects samples from the SO(3) sphere. Our method is designed to circumvent generating samples in regions unlikely to contain the true mode, thereby significantly improving data efficiency. As illustrated in \Cref{fig:diff_n_samples_perf}, \papername requires merely a few hundred samples to surpass the accuracy that RelPose++ achieves with 500,000 rotations. 
This ablation demonstrates \papername is sample efficient and its accuracy is not constraint by the resolution of the sampled grid. 

\textbf{Performance of the generator.}

\begin{figure}[]
\vspace{-4mm}
\begin{subfigure}{.49\linewidth}
\centering
\includegraphics[width=\linewidth]{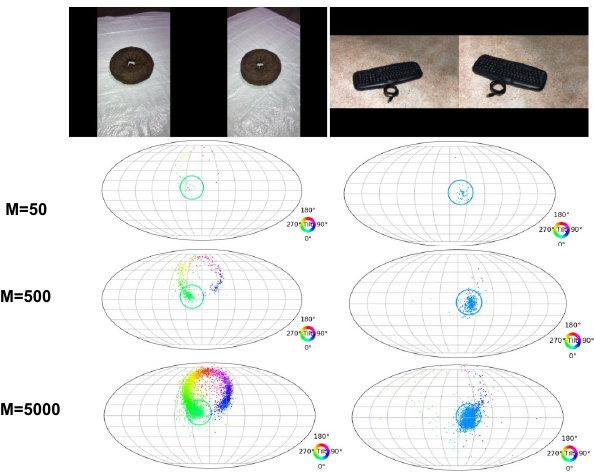}
\caption{Visualization of generator's output with different number of samples.}
\label{fig:vis_generator_num_sample}
\end{subfigure}
~
\begin{subfigure}{0.49\linewidth}
\centering
\includegraphics[width=\linewidth]{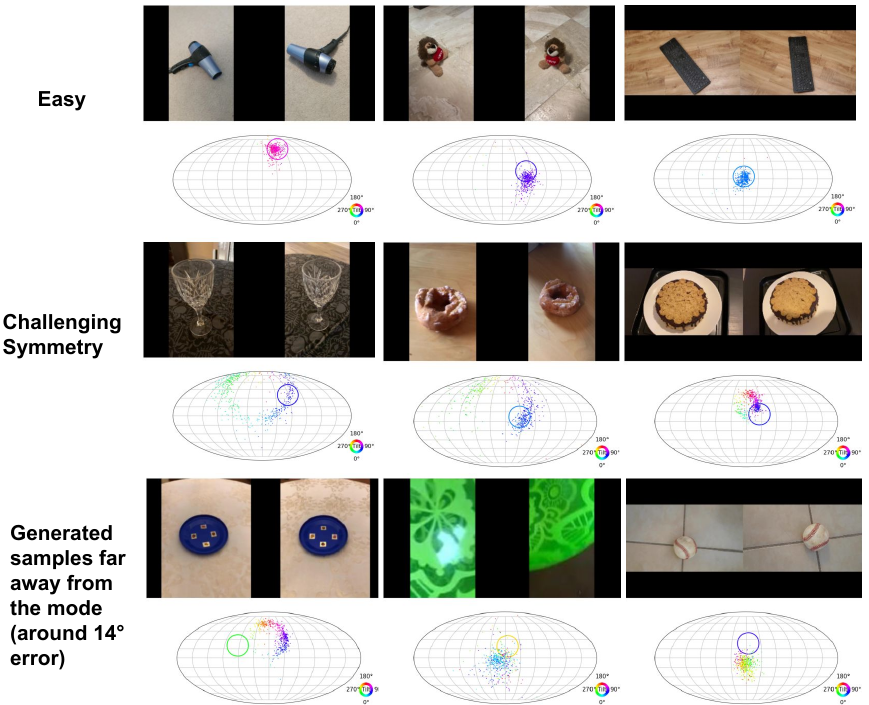}
\caption{Generator's output in challenging (symmetric) cases.}
\label{fig:vis_generator}
\end{subfigure}%

\caption{\textbf{Visualization of the output from generator.} Unfilled circles represent ground truth rotation. Dots represent generated rotation samples.}
\label{fig:generator_performance}
\vspace{-6mm}
\end{figure}

We present the following analysis to examine how the performance of the generator varies both qualitatively and quantitatively, given its critical role in overall performance. Specifically, \Cref{fig:generator_num_samples} illustrates the change in error for the optimal rotation hypothesis as the sample size is adjusted. Furthermore, \Cref{fig:vis_generator_num_sample} provides a visual representation of the rotation hypothesis distribution for varying numbers of samples (M). \Cref{fig:generator_performance} shows several examples of the generator's outputs for images that pose greater challenges, such as those with symmetry. Notably, there is an observed increase in the variance of hypotheses for these more difficult examples. However, the generator accurately models symmetry as a distribution, such as the donut and wine glass.


\subsection{Inference speed}
We also compared inference speeds across various methods, each evaluated using nine images as input, tested 50 times on an A6000 GPU. Since SparsePose is not open sourced, we directly quote their runtime from their paper. As can be seen from \Cref{table:inference_speed}, \papername reaches an average speed of 0.05s per 9 images and that is above 20 FPS, significantly outperforming all previous methods.
In summary, \papername not only surpasses state-of-the-art methods in accuracy but also demonstrates remarkable real-time speed, averaging 20 FPS with 9 input images, underscoring its efficiency and effectiveness in application.

\begin{table}[]
\vspace{-3mm}
\centering
\begin{tabular}{lcc}
\hline
                        & Time (secs)   & FPS         \\ \hline
COLMAP (SP + SG)        & 18            & 0.056       \\
SparsePose              & 3.6           & 0.278       \\
RelPose/RelPose++      & 48            & 0.020       \\
PoseDiff                & 63.8          & 0.015       \\
\textbf{Ours}           & \textbf{0.05} & \textbf{20}
\\ \hline
\end{tabular}
\caption{\textbf{Inference speed comparison}. \papername achieves significantly faster inference speed than other methods, averaging 20 FPS with nine images as input.  
}
\label{table:inference_speed}
\vspace{-12mm}
\end{table}


%% file: sections/conclusion.tex
\section{Conclusion}
\label{sec:conclusion}
In this paper, we present a new learning-based method for recovering camera poses from sparse-view RGB only images. The design of a pose generator and a pose discriminator in \papername empowers the network to navigate the ambiguity inherent in wide baseline images and generate multiple modes. Experiments demonstrates \papername achieves the SoTA performance on CO3D dataset, surpassing previous methods, particularly in accuracy at lower rotation error thresholds. Additionally, owing to its efficient generator, \papername can infer poses for nine images in real-time (20 FPS), demonstrating a significant speed improvement over all prior methods.